\documentclass{article}
\usepackage{graphicx}
\usepackage{amsmath,amssymb}
\usepackage[width=122mm,left=12mm,paperwidth=146mm,height=193mm,top=12mm,paperheight=217mm]{geometry}
\usepackage{url}
\usepackage{subfig}
\usepackage{wrapfig}

\begin{document}

\title{Pre-gen metrics: Predicting caption quality metrics without generating captions\footnote{This publication will appear in the Proceedings of the First Workshop on Shortcomings in Vision and Language (2018). DOI to be inserted later.}}

\author{
Marc Tanti \and Albert Gatt \and Adrian Muscat\\
University of Malta, Msida MSD 2080, Malta\\
\{marc.tanti.06, albert.gatt, adrian.muscat\}@um.edu.mt
}

\date{}

\maketitle

\begin{abstract}
Image caption generation systems are typically evaluated against reference outputs. We show that it is possible to predict output quality without generating the captions, based on the probability assigned by the neural model to the reference captions. Such pre-gen metrics are strongly correlated to standard evaluation metrics.
\end{abstract}

\section{Introduction}
%The evaluation of image description generation (IDG) systems relies heavily on automatic metrics, since collecting human judgments is an expensive process which is difficult to reproduce.
%The performance of iImage Description Generation (IDG) systems is best evaluated using human judgments, which invariably are  slow, expensive and difficult to reproduce. 
%It is therefore desirable to develop automatic metrics that correlate well with human judgments.
%, to help in speeding up the progress in IDG research.  
Automatic metrics for image description generation (IDG) compare \textit{c}, 
a generated caption,
%the sentence that has been generated by the  system, being compared 
to a set of reference sentences, ${R_1...R_n}$. 
We therefore refer to these as {\bf post-gen}(eration) metrics. 
%Many Image Description Generation (IDG) 
In most neural IDG architectures generation is performed by an algorithm such as beam search that samples the vocabulary at every timestep, selecting a likely next word after a given sentence prefix (according to the neural network) and attaching it to the end of the prefix, and repeating this procedure until the entire caption is produced. Given that the output thus generated is evaluated against a gold standard, post-gen metrics actually evaluate the neural network's ability to predict
%suggest 
the words in the reference captions given an image.
%during the generation process. 
Unfortunately, generating sentences is a time consuming process due to the fact that every word in a sentence requires its own forward pass through the neural network. This means that generating a 20-word sentence requires calling the neural network 20 times. 
As an indicative example, it takes 20.8 minutes to generate captions for every image in the MSCOCO test set on a standard hardware setup 
%%AG: SHould we include info RE CPU as well as GPU?
(GeForce GTX 760) using a beam width of just 1.
%, where the precise definition of similarity depends on the metric used. 

Our question is whether a system's performance can be assessed {\em prior} to the generation step, by exploiting the fact that the output is ultimately based on this core sampling mechanism. We envisage a scenario in which a neural caption generator is evaluated based on the extent to which its estimated 
%word probabilities 
softmax probabilities over the vocabulary 
%is compatible with 
maximise the probability of the words in the reference sentences ${R_1...R_n}$. We refer to this as a {\bf pre-gen}(eration) evaluation metric, as it can be computed prior to generating any captions. A well-known example of a pre-gen metric is language model perplexity although, as we show below, this metric is not the best pre-gen candidate in terms of its correlation to standard evaluation measures for IDG systems.

From a development perspective, the advantage of a pre-gen metric lies in that 
%it obviates the need to generate outputs to evaluate specific system configurations. While 
all the word probabilities in a sentence are immediately available to the network in one forward pass, whereas a post-gen metric can only be computed following a relatively expensive process of word-by-word generation requiring repeated calls to a neural network.
%In fact on our computer it only takes 28 seconds to compute a pre-gen metric like perplexity. 
To return to the earlier example, on the same hardware setup it only takes 28 seconds to compute model perplexity.

Thus, if pre-gen metrics can be shown to correlate strongly with established post-gen metrics, they could serve as a proxy for such metrics. This would speed up processes requiring repeated caption quality measurement such as during hyperparameter tuning. 

Finally, from a theoretical and empirical perspective, if caption quality, as measured by one or more post-gen metric(s), can be predicted prior to generation, this would shed further light on the underlying reasons for the observed correlations of such metrics \cite{kilickaya2017}. 

All code used in these experiments is publicly available.\footnote{See: \url{https://github.com/mtanti/pregen-metrics}}

The rest of this paper is organised as follows; background on metrics is covered in section 2, the methodology and experimental setup in section 3, and the results are given in section 4; the paper is concluded in section 5.
\section{Background: {\em Post-gen} metrics for image captioning}
%To compare the performance of image description generation (IDG) systems both human evaluations and automatic metrics have been used. 
%In the study of 
%The performance of IDG systems are best studied using human evaluations, which invariably are  slow, expensive and difficult to reproduce. It is therefore desirable to develop automatic metrics that correlate well with human judgments, to help in speeding up the progress in IDG research. 
%In this section we briefly review the most common metrics used to evaluate image description systems. 
In IDG, automatic metrics originally developed for Machine Translation or Summarisation, such as BLEU \cite{papineni2002}, ROUGE \cite{lin2004}, and METEOR \cite{banerjee2005}, were initially adopted, followed by metrics specifically designed for image description, notably CIDEr \cite{vedantam2015} and SPICE \cite{anderson2016}. Lately, Word Mover's Distance (WMD) \cite{kusner2015}, originally from the document similarity community, has also been suggested for IDG \cite{kilickaya2017}.
Like BLEU, ROUGE and METEOR, CIDEr 
%is a metric designed specifically for IDG.It 
makes use of n-gram similarities, while
%as the previous metric, 
%however it adds TF-IDF weights to the n-grams, and sums up the cosine similarity among n-grams to obtain the final score. 
WMD
%More recently Kilickaya et. al.\cite{kilickaya2017} proposed the use of WMD
%, originally developed as a document similarity metric, 
%for evaluating captions generated from an image. The WMD 
measures the semantic distance between texts on the basis of word2vec \cite{Mikolov2013} embeddings. %by minimizing the sum of the distance between individual words, computed using word2vec
%in the document. 
%The distance between words is computed from the word2vec 
%\cite{Mikolov2013} embeddings. 
%For image captioning the distance scores are converted to caption similarities using a negative exponential.
All of these
%the previously described 
metrics are purely linguistically informed.
%and none of them `looks' at the image during the evaluation process. 
By contrast, SPICE computes similarity between sentences from scene graphs \cite{Johnson2015}, obtained by parsing reference sentences.
This method is also linguistically informed; however the intuition behind it is that the human authored sentences should be an accurate reflection of image content. %are  based directly on the image.
%,
%to objects, attributes and relations, 
%which are intended as `proxies' for the image content, on the assumption that human-authored sentences
%. This method is also linguistically informed, as the others, however the intuition behind it is that the human authored sentences 
%are  based directly on the image.
%, (hence scene graph \cite{johnson:cvpr:2015}). 
%Finally, the SPICE metric is based on the matching of scene graph tuples.
%In summary, all these measures strive to compute the  similarity (in both lexical and semantic terms) between a set of sentences. A given image can be described with sentences that are not lexically similar but semantically similar. Therefore having a rich source of reference sentences often improves the performance of these metrics. 
%In summary, all the above metrics involve \textit{c}, the sentence that has been generated by the IDG system, being compared to a set of reference sentences, ${R_1...R_n}$, whether on lexical 
%purely string-based 
%or semantic grounds. 
%In practical terms, these metrics are all dependent on the system's having generated the output. We therefore refer to these as {\bf post-gen}(eration) metrics.
%\subsection{Relationship between metrics}

A typical IDG experiment reports several post-gen metrics. One reason is that the metrics correlate differently with human judgments, depending on task and dataset \cite{Bernardi2016}, echoing similar findings in other areas of NLP \cite{Dorr2004,Callison-Burch2006,Caporaso2008,Reiter2009,Cahill2009,Gatt2010,Espinosa2010,Wubben2012}. Thus, BLEU, METEOR, and ROUGE correlate weakly \cite{Kulkarni2011,Hodosh2013,Kiros2014} and yield different system rankings compared to human judgments \cite{vinyals2017}.  METEOR has a reportedly higher correlation than BLEU/ROUGE \cite{Elliott2013,Elliott2014}, with stronger relationships reported for CIDEr \cite{vedantam2015} and SPICE \cite{anderson2016}. 
Meta-evaluation of the ability of metrics to discriminate between captions have also been somewhat inconsistent \cite{vedantam2015,kilickaya2017}.

The extent to which post-gen metrics correlate with each other also varies, with stronger relationships among those based on n-grams on the one hand, and more semantically-oriented ones on the other \cite{kilickaya2017}, suggesting that these groups assess complementary aspects of quality, and partially explaining their variable relationship to human judgments in addition to variations due to dataset.

For neural IDG architectures, post-gen metrics have one fundamental property in common: they compare reference outputs to generated sentences which are based on sampling at each time-step from a probability distribution. Our hypothesis is that it is possible to exploit this, using the probability distribution itself to directly estimate the quality of captions, prior to generation.

\section{Pre-gen metrics}
Given a prefix, a neural caption generator predicts the next word by sampling from the softmax's probabilities estimated over the vocabulary. Let $R$ be a reference caption of length $m$. Given a prefix $R^{0\dots k}$ (where $R^0$ is the start token), $k \leq m$, a neural caption generator can be used to estimate the probability of the next word (or the end token) in the reference caption, $R^{k+1}$. The intuition underlying pre-gen metrics is that the higher the estimated probability of $R^{k+1}$, for all $k \leq m$, the more likely it is that the generator will approximate the reference caption. 
%%AG: Added the sentence below. Some people I talked to in the poster session weren't clear on this
Note that the idea is to estimate the probability of {\em reference} captions based on a trained IDG model.

\begin{figure}
	\centering
	\includegraphics[width=0.6\textwidth]{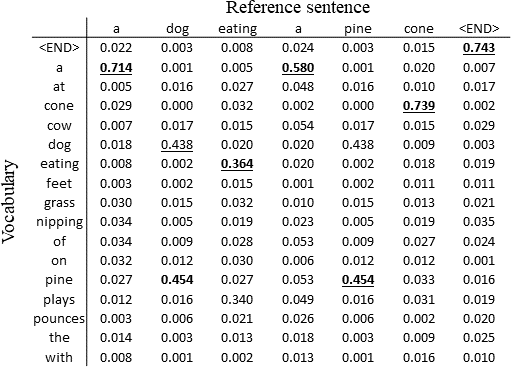}
	\caption{
		\small An example illustrating the output of a neural network that is predicting the probability for every word in a sentence. Given a single sentence, the caption generator will immediately output a matrix of probabilities, such that for every word position in the sentence, the matrix contains the probabilities for every word in the vocabulary being in that position given the image and the previous words (the first word has the start token as a previous word). In the above example, underlined probabilities are of the correct words being in the designated word position whilst bold probabilities are of the word with the maximum probability in the vocabulary for the designated word position. Note how the maximum probability is not always assigned to the correct word and that it might do so only intermittently.
		\label{fig:probs}
	}
\end{figure}

\begin{figure}[!h]
	\centering
	\includegraphics[width=0.9\textwidth]{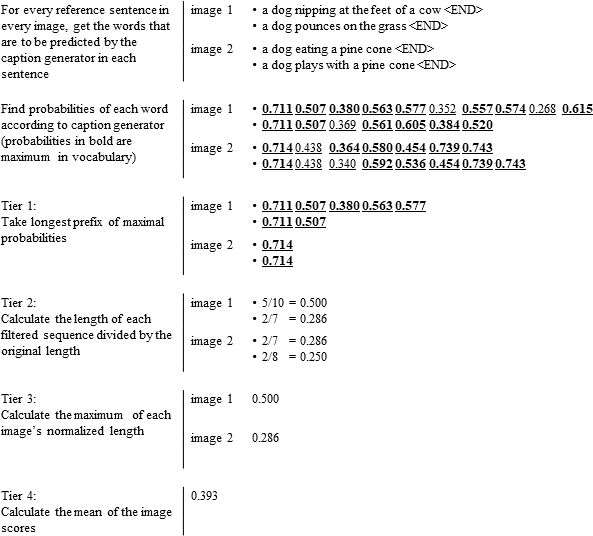}
	\caption{
		\small An example illustrating how tiers work. This illustration shows the best pre-gen metric found: \textit{mean}\textunderscore\textit{max}\textunderscore\textit{normcount}\textunderscore\textit{prefix0}.
		\label{fig:best-pregen}
	}
\end{figure}

Pre-gen metrics produce a score by aggregating the word probabilities predicted by the generator for all reference captions (combined with their respective image) over prefixes of different lengths. The way a caption generator predicts these word probabilities is illustrated in Figure~\ref{fig:probs}. To find the best way to aggregate the word probabilities, we define a search space by setting options at four different algorithmic steps which we call `tiers'. Each tier represents a function and the composition of all four tiers together constitutes a pre-gen function. We try several different options for each tier in order to find the best pre-gen function. Figure~\ref{fig:best-pregen} shows an example of how tiers form a pre-gen function.

Given a set of images with their corresponding reference captions, the process starts by computing each reference caption's individual word probabilities (given the image) according to the model. Note that the model may not predict every word in a reference caption as the most likely in the vocabulary.

The first tier is a filter that selects which predicted word probabilities should be considered in the next tier. %The intuition is to only take into consideration words in the ground truth sentence that are likely to be generated. 
We consider three possible filters: (a) the filter \textit{none} passes all probabilities; (b) \textit{filter0} filters out the word probabilities that are not ranked as most probable in the vocabulary by the model, i.e are not predicted to be maximally probable continuations of the current prefix; and (c) \textit{prefix0} selects the longest prefix of the caption such that the model predicts all words in the prefix as being the most likely in the vocabulary.
%passes the longest prefix that consists of the uninterrupted list of maximum probabilities.

At the second tier, we aggregate the selected word probabilities in each reference sentence into a single score for each sentence. We define four possible functions: (a) \textit{prob} multiplies all probabilities; (b) \textit{pplx} computes the perplexity; (c) \textit{count} counts the number of word probabilities that were selected in the first tier; and (d) \textit{normcount} normalises \textit{count} by the total number of words in the reference sentence.

The third tier aggregates the scores obtained for all reference sentences into a single score for each image. We explore six possibilities: (a) \textit{sum}; (b) \textit{mean}; (c) \textit{median}; (d) \textit{geomean}, the geometric mean; (e) \textit{max}; and (f) \textit{min}. We also consider (g) \textit{join}, whereby all the image-sentence scores are joined into a single list without aggregation so that they are all aggregated together in the next tier.

%The third tier aggregates the sentence scores for each image into a single score per image. This can be either the \textit{sum}, \textit{mean}, \textit{median}, geometric mean (\textit{geomean}), maximum (\textit{max}), or minimum (\textit{min}). We also consider joining all the image-sentence scores into a single list of scores without aggregating anything so that they are all aggregated together in the next tier (\textit{join}).

The fourth tier aggregates the image scores into a single dataset score, which is the final pre-gen score of the caption generator. For this aggregation, we use the same functions in the previous tier excluding \textit{join}.
%We define the same six function as in tier three, i.e. (a) \textit{sum}, (b) \textit{mean}, (c) \textit{median}, (d) \textit{geomean}, (e) \textit{max}, and (f) \textit{min}.

The above possibilities result in 504 unique combinations. In what follows, we adopt the convention of denoting a pre-gen metric by the sequence of function names that compose it, starting from tier four e.g. \textit{mean}\textunderscore\textit{max}\textunderscore\textit{normcount}\textunderscore\textit{prefix0}. In our experiments, we compute all of these different combinations and compare their predictions to standard post-gen metrics, namely METEOR, CIDEr, SPICE, and WMD. All metrics except WMD were computed using the MSCOCO Evaluation toolkit\footnote{See: \url{https://github.com/tylin/coco-caption}}. Since the toolkit does not include WMD, we created a fork of the repository that includes it.\footnote{See: \url{https://github.com/mtanti/coco-caption}}
%We focus our attention on the metrics with the highest correlations.
%The next section gives the results.

\subsection{Experimental setup}
%To generate the word probabilities for computing the pre-gen metrics and to compute 
For our experiments, we used a variety of pre-trained neural caption generators (36 in all) from
\cite{Tanti2018}.\footnote{See: \url{https://github.com/mtanti/where-image2}} These models are based on four different caption generator architectures. Each was trained and tested over three runs on Flickr8K \cite{Hodosh2013}, Flickr30k \cite{Young2014}, and MSCOCO \cite{Lin2014}. The four architectures differ in terms of how the CNN image encoder is combined with the RNN: {\tt init} architectures use the image vector as the initial hidden state of the RNN; {\tt pre} architectures treat the image vector as the first word of a caption; {\tt par} architectures are trained on captions where each word is coupled with an image vector at each time-step; and {\tt merge} architectures keep the image out of the the RNN entirely, merging the image vector with the RNN hidden state in a final feedforward layer, prior to prediction.

\begin{figure}
	\centering
	\includegraphics[width=0.6\textwidth]{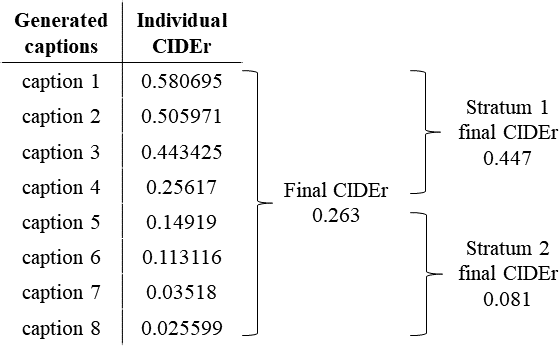}
	\caption{
		\small An example illustrating how a dataset is broken into strata in order to create a variety of performance scores using the same neural networks and thus be able to test the correlation between the post-gen and pre-gen metrics on many different scores.
		\label{fig:strata}
	}
\end{figure}

Since only the final trained versions of the models are available, there is a bias towards good quality post-gen metric results. This renders the values of the post-gen metrics rather similar and concentrated in a small range. 
%It is desired to have a useful pre-gen metric be one that works equally well on bad models as well as good ones. 
A pre-gen metric is useful if it makes good predictions on models of any quality not just good ones.
Rather than re-training all the models and saving the parameters at different intervals during training, we opted to  stratify the dataset on the basis of how well each individual image is rated by the CIDEr metric. This is illustrated in Figure~\ref{fig:strata}.

%A single score is normally computed to quantify the quality of all the generated captions. Alternatively, individual scores can be calculated for each individual generated caption. 
We grouped images into the best and worst halves on the basis of the CIDEr score (since CIDEr is the post-gen metric that best correlates with the other post-gen metrics \cite{kilickaya2017}) of their sentences as generated by a model. This creates two datasets, one where the model performs well and one where the model performs badly.
%, such that we obtain a higher and lower final score for each half. 
We stratified the dataset into different numbers of equal parts and not just two, namely: 1 (whole), 2, 3, 4 and 5, resulting in a 15-fold increase over the original 36 averaged results and more importantly, over a wide dynamic range in CIDEr scores, 
which we required to study the correlation in between pre- and post-gen metrics.
\section{Results}
% Figure \ref{fig:post-gen} shows the mean post-gen metric scores for each of the four architectures on the three datasets.

% \begin{figure}[!h]
% \centering
% % \includegraphics[width=\textwidth,height=0.3\textheight]{img/postgen-score-means}
% \includegraphics[scale=0.45]{img/postgen-score-means}
% \caption{\small Post-gen metrics. Error bars represent $\pm$ 1 standard error of the mean.}
% \label{fig:post-gen}
% \end{figure}

We evaluate the correlation between pre- and post-gen metrics using the Coefficient of Determination, or $R^2$, defined as the square of the Pearson correlation coefficient. The reason for this is twofold. First, $R^2$ reflects the magnitude of a correlation, irrespective of whether it is positive or negative (the pre-gen metrics based on perplexity would be expected to be negatively correlated with post-gen metrics). Second, given a linear model in which a pre-gen metric is used to predict the value on a post-gen metric, $R^2$ indicates the proportion of the variance in the latter that the pre-gen metric predicts.

\begin{figure}[!h]
\centering
\subfloat[{\sc cider}] {
       \includegraphics[width=0.4\textwidth]{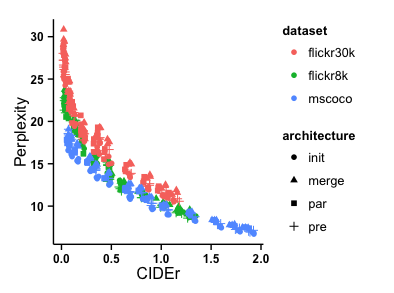}
        \label{fig:cider-pplx}
}
\qquad
\subfloat[{\sc spice}]{
       \includegraphics[width=0.4\textwidth]{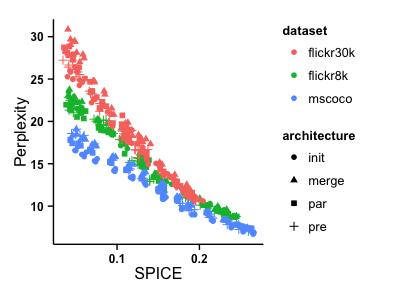}
        \label{fig:spice-pplx}
       }
       
\subfloat[{\sc meteor}]{
       \includegraphics[width=0.4\textwidth]{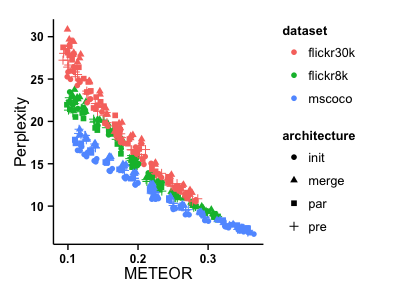}
        \label{fig:meteor-pplx}
       }    
\qquad       
\subfloat[{\sc wmd}]{
       \includegraphics[width=0.4\textwidth]{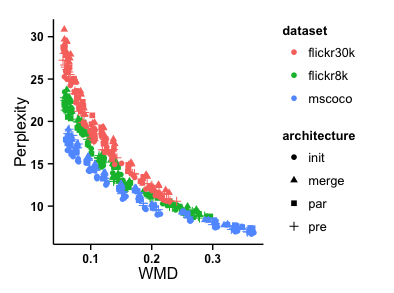}
        \label{fig:wmd-pplx}
       }       
\caption{\small Relationship between perplexity and post-gen metrics by dataset and architecture. The overall correlation has an $R^2$ of 0.76. (Best viewed in colour.)}
\label{fig:pplx-by-postgen}
\end{figure}

As a baseline, we show the scatter plot for the relationship between language model perplexity and the post-gen metrics in Figure~\ref{fig:pplx-by-postgen}. In terms of the description in the previous section, perplexity is defined as \textit{geomean}\textunderscore\textit{join}\textunderscore\textit{pplx}\textunderscore\textit{none}. As can be seen, perplexity performs somewhat poorly on low scoring captions. Our question is whether a better pre-gen metric can be found.

For each of the 4 post-gen metrics, we identified the top 5 best correlated pre-gen metrics, based on the $R^2$ value computed over all the data (i.e. aggregating scores across architectures and datasets). The top 4 pre-gen metrics were the same for all post-gen metrics, namely: 
\begin{enumerate}
\item \textit{mean}\textunderscore\textit{max}\textunderscore\textit{normcount}\textunderscore\textit{prefix0}; 
\item \textit{mean}\textunderscore\textit{mean}\textunderscore\textit{normcount}\textunderscore\textit{prefix0}; 
\item \textit{mean}\textunderscore\textit{join}\textunderscore\textit{normcount}\textunderscore\textit{prefix0}; 
\item \textit{mean}\textunderscore\textit{sum}\textunderscore\textit{normcount}\textunderscore\textit{prefix0}
\end{enumerate}

Note that all the best performing metrics are based on the variable \textit{prefix0}. This is not surprising since when generating a sentence, it is probably the word with the maximum probability in the vocabulary which gets selected as a next word in a prefix. Hence if the most probable next word is an incorrect one then it will likely send the rest of the caption generation process off the rails as it will misinform the next words as well. Hence, \textit{prefix0} is a measure of how likely this is to happen.

On the other hand, the fifth most highly correlated pre-gen metric differed for each post-gen metric, as follows:

\begin{itemize}
\item CIDER: \textit{mean}\textunderscore\textit{min}\textunderscore\textit{count}\textunderscore\textit{filter0};
\item METEOR: \textit{mean}\textunderscore\textit{mean}\textunderscore\textit{count}\textunderscore\textit{prefix0};
\item SPICE: \textit{geomean}\textunderscore\textit{min}\textunderscore\textit{pplx}\textunderscore\textit{filter0};
\item WMD: \textit{mean}\textunderscore\textit{join}\textunderscore\textit{count}\textunderscore\textit{prefix0}
\end{itemize}

\begin{figure}[!t]
\centering
\includegraphics[width=\textwidth]{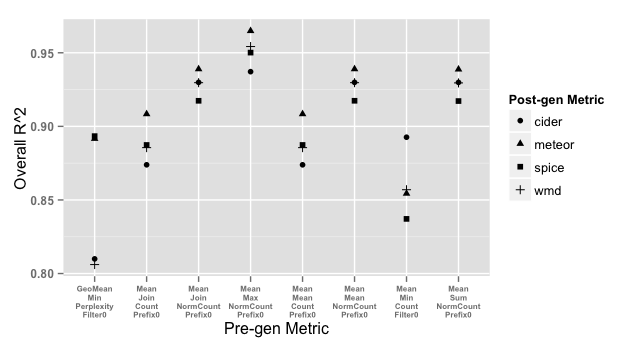}
\caption{\small Overall $R^2$ between the 4 post-gen metrics and their 5 most highly correlated pre-gen metrics. Scores average over architectures and datasets.}
\label{fig:top5-pregen}
\end{figure}

Figure \ref{fig:top5-pregen} displays the correlation between these pre-gen metrics and the post-gen scores. Note that all $R^2$ scores are above 0.8, indicating a very strong correlation.\footnote{All correlations are significant at $p < 0.001$.} The top 4 scores have $R^2 \geq 0.9$. 

\begin{figure}[!t]
\centering
\subfloat[{\sc cide}r] {
       \includegraphics[width=0.4\textwidth]{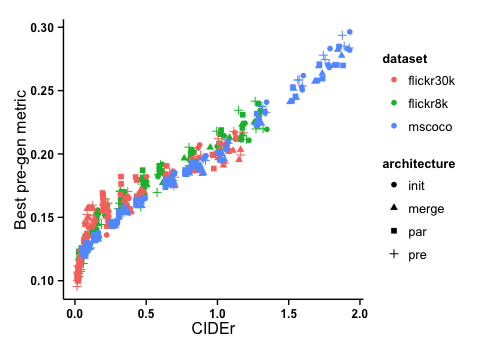}
        \label{fig:cider-pregen}
}
\qquad
\subfloat[{\sc spice}]{
       \includegraphics[width=0.4\textwidth]{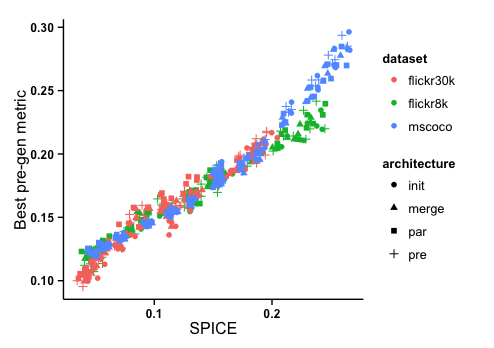}
        \label{fig:spice-pregen}
       }
       
\subfloat[{\sc meteor}]{
       \includegraphics[width=0.4\textwidth]{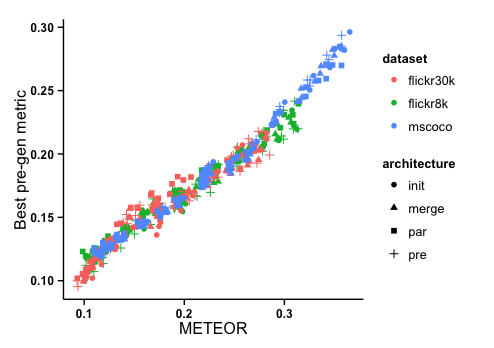}
        \label{fig:meteor-pregen}
       }       
\qquad       
\subfloat[{\sc wmd}]{
       \includegraphics[width=0.4\textwidth]{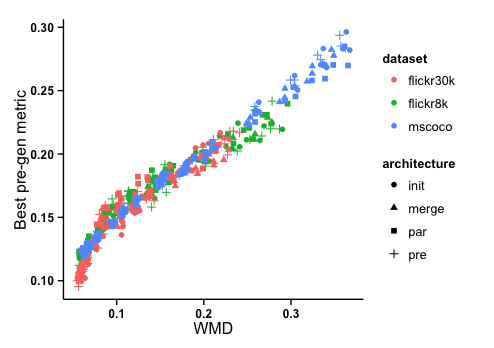}
        \label{fig:wmd-pregen}
       }       
\caption{\small Best pre-gen metric (\textit{mean}\textunderscore\textit{max}\textunderscore\textit{normcount}\textunderscore\textit{prefix0}) vs post-gen metrics. The overall correlation has an $R^2$ of 0.94. (Best viewed in colour.)
}
\label{fig:best-pregen-by-postgen}
\end{figure}

To investigate the relationship between pre- and post-gen metrics more closely, we focus on the best pre-gen metric (that is, \textit{mean}\textunderscore\textit{max}\textunderscore\textit{normcount}\textunderscore\textit{prefix0}) and consider its relationship to each post-gen metric individually. This is shown in Figure \ref{fig:best-pregen-by-postgen}. Irrespective of architecture and/or dataset, we observe a broadly linear relationship, despite some evidence of non-linearity at the lower ends of the scale, especially for CIDEr and WMD. This supports the hypothesis made at the outset, namely, that it is possible to predict the quality of captions, as measured by a standard metric, by considering the probability of the reference captions in the test set, without the need to generate the captions themselves.

%\section{Discussion}
%This is sensible because when generating a sentence, it is probably the word with the maximum probability in the vocabulary which gets selected as a next word in a prefix. What's more is that it is probably the case that as soon as one wrong word is used in a sentence, the remainder of the sentence will also be wrong. These filters are used to only take into consideration words in the ground truth sentence that are likely to be generated.

\section{Discussion and conclusion}
We have introduced and defined the concept of pre-gen metrics and described a methodology to search for useful variants of these metrics. Our results show that pre-gen metrics closely approximate a variety of standard evaluation measures. 

These results can be attributed to the fact that neural captioning models share core assumptions about the sampling mechanisms that underlie generation, and that standard evaluation metrics ultimately assess the output of this sampling process. Thus, it is possible to predict the quality of the output, as measured by a post-gen metric, using the probability distribution that a trained model predicts over prefixes of varying length in the reference captions. The practical implication is that pre-gen metrics can act as quick and efficient evaluation proxies during development. The theoretical implication is that the correlations among standard evaluation metrics reported in the literature are due, at least in part, to core sampling mechanisms shared by most neural generation architectures.

In future work, we plan to experiment with tuning captioning models using pre-gen metrics. We also wish to compare pre-gen metrics directly to human judgments.

%This work was motivated by the observation that neural caption generators rely on a core sampling mechanism, and that standard evaluation metrics ultimately assess the quality of outputs that arise from this sampling mechanism. Our results support this hypothesis. 
%Additionally, it is probably the case that the selection of an \textit{unsuitable} word throws the generation process off the reference sentences. As future work, we will tune the models using our pre-gen metrics and report on the results.

\section*{Acknowledgments}
{\small The research in this paper is partially funded by the Endeavour Scholarship Scheme (Malta). Scholarships are part-financed by the European Union - European Social Fund (ESF) - Operational Programme II – Cohesion Policy 2014-2020 “Investing in human capital to create more opportunities and promote the well-being of society”.}

\clearpage

\bibliographystyle{plain}
\bibliography{bibliography}

\begin{thebibliography}{10}

\bibitem{anderson2016}
Peter Anderson, Basura Fernando, Mark Johnson, and Stephen Gould.
\newblock Spice: Semantic propositional image caption evaluation.
\newblock In {\em ECCV}, 2016.

\bibitem{banerjee2005}
Satanjeev Banerjee and Alon Lavie.
\newblock {METEOR}: {An} automatic metric for {MT} evaluation with improved
  correlation with human judgments.
\newblock In {\em Proc. Workshop on Intrinsic and extrinsic evaluation measures
  for machine translation and/or summarization}, volume~29, pages 65--72, 2005.

\bibitem{Bernardi2016}
Raffaella Bernardi, Ruket Cakici, Desmond Elliott, Aykut Erdem, Erkut Erdem,
  Nazli Ikizler-Cinbis, Frank Keller, Adrian Muscat, and Barbara Plank.
\newblock {Automatic} {Description} {Generation} from {Images}: {A} {Survey} of
  {Models}, {Datasets}, and {Evaluation} {Measures}.
\newblock {\em JAIR}, 55:409--442, 2016.

\bibitem{Cahill2009}
Aoife Cahill.
\newblock {Correlating Human and Automatic Evaluation of a German Surface
  Realiser}.
\newblock In {\em Proc. ACL-IJCNLP'09}, pages 97--100, 2009.

\bibitem{Callison-Burch2006}
Chris Callison-Burch, Miles Osborne, and Philipp Koehn.
\newblock {Re-evaluating the Role of BLEU in Machine Translation Research}.
\newblock In {\em Proc. EACL'06}, pages 249--256, 2006.

\bibitem{Caporaso2008}
J.~Gregory Caporaso, Nita Deshpande, J.~Lynn Fink, Philip~E. Bourne, Kevin
  Bretonnel~Cohen, and Lawrence Hunter.
\newblock {Intrinsic evaluation of text mining tools may not predict
  performance on realistic tasks}.
\newblock {\em Pacific Symposium on Biocomputing}, 13:640--651, 2008.

\bibitem{Dorr2004}
B~Dorr, Christof Monz, Douglas Oard, Stacy President, David Zajic, and Richard
  Schwartz.
\newblock {Extrinsic Evaluation of Automatic Metrics}.
\newblock Technical report, Instititue for Advanced Computer Studies, Univ of
  Maryland, College Park, College Park, MD, 2004.

\bibitem{Elliott2013}
Desmond Elliott and Frank Keller.
\newblock Image description using visual dependency representations.
\newblock In {\em Proceedings of the 2013 Conference on Empirical Methods in
  Natural Language Processing}, pages 1292--1302, Seattle, Washington, USA,
  October 2013. Association for Computational Linguistics.

\bibitem{Elliott2014}
Desmond Elliott and Frank Keller.
\newblock {Comparing Automatic Evaluation Measures for Image Description}.
\newblock In {\em Proc. ACL'14}, pages 452--457, 2014.

\bibitem{Espinosa2010}
Dominic Espinosa, Rajakrishnan Rajkumar, Michael White, and Shoshana Berleant.
\newblock {Further Meta-Evaluation of Broad-Coverage Surface Realization}.
\newblock In {\em Proc. EMNLP'10}, pages 564--574, 2010.

\bibitem{Gatt2010}
Albert Gatt and Anja Belz.
\newblock {Introducing shared task evaluation to NLG: The TUNA shared task
  evaluation challenges}.
\newblock In Emiel Krahmer and Mari\"{e}t Theune, editors, {\em Empirical
  methods in natural language generation}. Springer, Berlin and Heidelberg,
  2010.

\bibitem{Hodosh2013}
Micah Hodosh, Peter Young, and Julia Hockenmaier.
\newblock {Framing} {Image} {Description} as a {Ranking} {Task}: {Data},
  {Models} and {Evaluation} {Metrics}.
\newblock {\em JAIR}, 47(1):853--899, 2013.

\bibitem{Johnson2015}
Justin Johnson, Ranjay Krishna, Michael Stark, Li-Jia Li, David~A. Shamma,
  Michael~S. Bernstein, and Li~Fei-Fei.
\newblock Image retrieval using scene graphs.
\newblock In {\em 2015 {IEEE} Conference on Computer Vision and Pattern
  Recognition ({CVPR})}. {IEEE}, June 2015.

\bibitem{kilickaya2017}
Mert Kilickaya, Aykut Erdem, Nazli Ikizler-Cinbis, and Erkut Erdem.
\newblock Re-evaluating automatic metrics for image captioning.
\newblock In {\em Proceedings of the 15\textsuperscript{th} Conference of the
  European Chapter of the Association for Computational Linguistics: Volume 1,
  Long Papers}. Association for Computational Linguistics, 2017.

\bibitem{Kiros2014}
Ryan Kiros, Ruslan Salakhutdinov, and Richard~S. Zemel.
\newblock {Unifying} visual-semantic embeddings with multimodal neural language
  models.
\newblock {\em CoRR}, 1411.2539, 2014.

\bibitem{Kulkarni2011}
Girish Kulkarni, Visruth Premraj, Sagnik Dhar, Siming Li, Yejin Choi,
  Alexander~C. Berg, and Tamara~L. Berg.
\newblock Baby talk: Understanding and generating simple image descriptions.
\newblock In {\em {CVPR} 2011}. {IEEE}, June 2011.

\bibitem{kusner2015}
Matt Kusner, Yu~Sun, Nicholas Kolkin, and Kilian Weinberger.
\newblock From word embeddings to document distances.
\newblock In Francis Bach and David Blei, editors, {\em Proceedings of the
  32\textsuperscript{nd} International Conference on Machine Learning},
  volume~37 of {\em Proceedings of Machine Learning Research}, pages 957--966,
  Lille, France, 07--09 Jul 2015. PMLR.

\bibitem{lin2004}
Chin-Yew Lin and Franz~Josef Och.
\newblock Automatic evaluation of machine translation quality using longest
  common subsequence and skip-bigram statistics.
\newblock In {\em Proc. {ACL}'04}, 2004.

\bibitem{Lin2014}
Tsung-Yi Lin, Michael Maire, Serge Belongie, James Hays, Pietro Perona, Deva
  Ramanan, Piotr Doll{\'{a}}r, and C.~Lawrence Zitnick.
\newblock Microsoft {COCO}: Common objects in context.
\newblock In {\em Proc. ECCV'14}, pages 740--755, 2014.

\bibitem{Mikolov2013}
Tomas Mikolov, Kai Chen, Greg Corrado, and Jeffrey Dean.
\newblock {Efficient} {Estimation} of {Word} {Representations} in {Vector}
  {Space}.
\newblock {\em CoRR}, 1301.3781, 2013.

\bibitem{papineni2002}
Kishore Papineni, Salim Roukos, Todd Ward, and Wei-Jing Zhu.
\newblock {BLEU}: {A} method for automatic evaluation of machine translation.
\newblock In {\em Proc. ACL'02}, pages 311--318, 2002.

\bibitem{Reiter2009}
Ehud Reiter and Anja Belz.
\newblock {An Investigation into the Validity of Some Metrics for Automatically
  Evaluating Natural Language Generation Systems}.
\newblock {\em Computational Linguistcs}, 35(4):529--558, 2009.

\bibitem{Tanti2018}
Marc Tanti, Albert Gatt, and Kenneth~P. Camilleri.
\newblock Where to put the image in an image caption generator.
\newblock {\em Natural Language Engineering}, 24(3):467--489, April 2018.

\bibitem{vedantam2015}
Ramakrishna Vedantam, C.~Lawrence Zitnick, and Devi Parikh.
\newblock {CIDEr}: Consensus-based image description evaluation.
\newblock In {\em Proc. CVPR'15}, 2015.

\bibitem{vinyals2017}
Oriol Vinyals, Alexander Toshev, Samy Bengio, and Dumitru Erhan.
\newblock Show and tell: Lessons learned from the 2015 {MSCOCO} image
  captioning challenge.
\newblock {\em {IEEE} Transactions on Pattern Analysis and Machine
  Intelligence}, 39(4):652--663, April 2017.

\bibitem{Wubben2012}
Sander Wubben, Antal van~den Bosch, and Emiel Krahmer.
\newblock {Sentence Simplification by Monolingual Machine Translation}.
\newblock In {\em Proc. ACL'12}, pages 1015--1024, 2012.

\bibitem{Young2014}
Peter Young, Alice Lai, Micah Hodosh, and Julia Hockenmaier.
\newblock From image descriptions to visual denotations: {New} similarity
  metrics for semantic inference over event descriptions.
\newblock {\em TACL}, 2:67--78, 2014.

\end{thebibliography}
\end{document}